\documentclass[11pt]{article}

\usepackage[margin=1in]{geometry}
\usepackage{amsmath,amssymb}
\usepackage{booktabs}
\usepackage{pgfplots}
\usepackage{tikz}
\usepackage{hyperref}
\usepackage{natbib}
\usepackage{xcolor}
\usepackage{enumitem}
\usepackage{microtype}
\usepackage{algorithm}
\usepackage{algpseudocode}
\usepackage{subcaption}

\pgfplotsset{compat=1.18}
\usetikzlibrary{positioning,arrows.meta,shapes.geometric,fit,calc}

\definecolor{aneblue}{HTML}{2563EB}
\definecolor{cpuorange}{HTML}{EA580C}
\definecolor{nanred}{HTML}{DC2626}
\definecolor{fixgreen}{HTML}{16A34A}
\definecolor{layergray}{HTML}{F1F5F9}

\hypersetup{
  colorlinks=true,
  linkcolor=aneblue,
  citecolor=aneblue,
  urlcolor=aneblue
}

\title{\textbf{Orion: Characterizing and Programming Apple's Neural Engine\\for LLM Training and Inference}}

\author{
  Ramchand Kumaresan\\
  \texttt{https://github.com/mechramc/Orion}
}

\date{}

\begin{document}
\maketitle

\begin{abstract}
Over two billion Apple devices ship with a Neural Processing Unit (NPU) --- the Apple Neural Engine (ANE) --- yet this accelerator remains almost entirely unused for large language model workloads. CoreML, Apple's public ML framework, imposes opaque abstractions that prevent direct ANE programming and do not support on-device training. We present \textsc{Orion}, to our knowledge the first open end-to-end system that combines direct ANE execution, a compiler pipeline, and stable multi-step training with checkpoint resume in a single native runtime, bypassing CoreML entirely via Apple's private \texttt{\_ANEClient} and \texttt{\_ANECompiler} APIs. Building on prior characterization work by maderix, who reverse-engineered the private API surface and benchmarked ANE hardware characteristics, we extend the public knowledge of ANE constraints to a catalog of 20 restrictions on MIL IR programs, memory layout, compilation limits, and numerical behavior --- including 14 previously undocumented constraints discovered during \textsc{Orion} development. \textsc{Orion} includes a compiler that lowers a graph IR through five optimization passes to ANE-native MIL, and a runtime that manages IOSurface-backed zero-copy tensor I/O, program caching, and delta compilation for weight updates. A key contribution is \emph{delta compilation}: since the ANE bakes weights at compile time, na\"ive training requires full recompilation per step ($\sim$4.2\,s). We show that compiled programs can be surgically updated by unloading, patching the weight files on disk, and reloading --- bypassing \texttt{ANECCompile()} entirely and reducing recompilation from 4,200\,ms to 494\,ms per step (8.5$\times$), yielding a 3.8$\times$ total training speedup. On an M4 Max, \textsc{Orion} achieves 170+ tokens/s for GPT-2 124M inference and demonstrates stable training of a 110M-parameter transformer on TinyStories for 1,000 steps in 22 minutes with zero NaN occurrences. We also present LoRA adapter-as-input, enabling hot-swap of low-rank adapters via IOSurface inputs without recompilation. We release \textsc{Orion} as open source under the MIT license.
\end{abstract}

\section{Introduction}

Apple's Neural Engine (ANE) is a dedicated neural processing unit present in every Apple silicon chip since the A11 Bionic (2017). With the M4 generation, the ANE delivers up to 38 TOPS (INT8) across 16 cores, rivaling discrete AI accelerators in raw throughput. Over two billion active Apple devices carry some variant of this hardware~\citep{coreml2023}. Yet despite this enormous installed base, the ANE remains a \emph{dark accelerator} for large language models: no public framework supports LLM training on ANE, and inference frameworks universally target the GPU via Metal or the CPU.

The root cause is Apple's software stack. CoreML~\citep{coreml2023}, the only public interface to the ANE, operates as a black-box scheduler that decides at runtime whether to dispatch operations to the CPU, GPU, or ANE. Developers cannot force ANE execution, inspect ANE programs, or perform gradient computation. The ANE's native instruction set --- compiled from Apple's Model Intermediate Language (MIL) --- is undocumented, and the compilation and evaluation APIs reside in a private framework (\texttt{AppleNeuralEngine.framework}).

Prior work has begun to crack this barrier. The maderix project~\citep{maderix2024ane,maderix2024ane_part1,maderix2024ane_part2} made the foundational breakthrough: reverse-engineering the private API calling sequence, demonstrating direct ANE dispatch from C, and producing the first empirical characterization of ANE hardware --- including debunking Apple's 38 TOPS specification (actual fp16 throughput: $\sim$19 TFLOPS), discovering the 32 MB SRAM performance cliff, measuring dispatch overhead ($\sim$0.095 ms), and identifying the $\sim$119 compilation-per-process limit. ANEgpt~\citep{anegpt2024} extended this to transformer training, implementing forward and backward passes for a 110M-parameter model. Hollemans~\citep{hollance2022ane} documented ANE characteristics through CoreML-level experiments, establishing the preference for 1$\times$1 convolutions and \texttt{[B, C, 1, S]} tensor layouts. However, none of these efforts produced a complete system: ANEgpt could not resume training without NaN divergence, lacked a compiler, and used Python for orchestration; maderix's characterization focused on hardware-level benchmarking without building an LLM runtime.

We present \textsc{Orion}, which to our knowledge is the first open system to combine direct ANE execution with a compiler pipeline, reproducible inference, and stable multi-step training in a single native runtime. Our contributions are:

\begin{enumerate}[leftmargin=*,itemsep=2pt]
  \item \textbf{ANE Characterization.} A consolidated catalog of 20 ANE programming constraints (Section~\ref{sec:characterization}), extending prior hardware-level characterization by maderix~\citep{maderix2024ane_part1,maderix2024ane_part2} with 14 newly discovered MIL IR restrictions, memory layout requirements, and numerical behaviors found during \textsc{Orion} development.

  \item \textbf{Compiler.} A graph IR with 27 operations lowered through five optimization passes (DCE, identity elimination, cast fusion, SRAM annotation, constraint validation) to ANE-native MIL, with 13 verified frontends plus LoRA-fused variants (Section~\ref{sec:system}).

  \item \textbf{Delta Compilation.} A technique that bypasses \texttt{ANECCompile()} for weight updates by unloading compiled programs, patching weight files on disk, and reloading --- reducing recompilation overhead from 4,200\,ms to 494\,ms (8.5$\times$) and eliminating the $\sim$119 compile-per-process limit entirely (Section~\ref{sec:delta}).

  \item \textbf{Training on ANE.} Stable multi-step training of a 110M-parameter transformer with automatic checkpoint resume, achieved by solving three NaN-inducing bugs through deferred compilation, fp16 overflow clamping, and gradient sanitization (Section~\ref{sec:numerical}).

  \item \textbf{LoRA Adapter-as-Input.} Low-rank adapter matrices passed as IOSurface inputs rather than baked weights, enabling hot-swap of adapters without recompilation (Section~\ref{sec:lora}).

  \item \textbf{Open-source release.} A complete Objective-C runtime --- Python is used only for one-time weight conversion from HuggingFace formats; inference, training, and benchmarking require no Python --- with GPT-2 124M inference (170+ tok/s), Stories110M training (1,000 steps in 22 minutes), native BPE and SentencePiece tokenizers, released under the MIT license.
\end{enumerate}

\section{Background}
\label{sec:background}

\subsection{Apple Neural Engine Hardware}

The ANE is a fixed-function accelerator optimized for convolution and matrix-multiply workloads in fp16 precision. Table~\ref{tab:ane_hw} summarizes the hardware characteristics of the M4 Max generation, drawing on measurements by maderix~\citep{maderix2024ane_part2} and confirmed independently in our experiments.

\begin{table}[h]
\centering
\caption{Apple Neural Engine hardware characteristics (M4 Max).}
\label{tab:ane_hw}
\begin{tabular}{ll}
\toprule
\textbf{Property} & \textbf{Value} \\
\midrule
Generation & H16 \\
Neural Engine cores & 16 \\
Peak throughput (INT8, Apple spec) & 38 TOPS \\
Peak throughput (fp16, measured\footnotemark) & $\sim$19 TFLOPS \\
On-chip SRAM\footnotemark & 32 MB \\
Evaluation queue depth & 127 \\
Dispatch overhead (XPC+IOKit) & $\sim$0.095 ms \\
Idle power & Zero (hard power gating) \\
\bottomrule
\end{tabular}
\end{table}

\addtocounter{footnote}{-1}
\stepcounter{footnote}\footnotetext{Apple specifies 38 TOPS (INT8), but the ANE dequantizes INT8 to fp16 before computation. Actual measured peak is $\sim$19 TFLOPS fp16~\citep{maderix2024ane_part2}; INT8 saves only memory bandwidth, not compute cycles.}
\stepcounter{footnote}\footnotetext{Performance drops $\sim$30\% when working sets exceed the 32 MB SRAM budget, forcing spills to DRAM~\citep{maderix2024ane_part2}.}

The ANE operates on a \emph{compile-then-dispatch} model. Programs are expressed in Apple's Model Intermediate Language (MIL), compiled to E5 microcode by \texttt{\_ANECompiler}, and evaluated via \texttt{\_ANEClient}. All tensor I/O uses IOSurface-backed shared memory in a fixed \texttt{[1, C, 1, S]} layout (fp16), enabling zero-copy data transfer between the CPU and ANE.

Critically, the ANE \emph{bakes weights at compile time}: weight tensors are embedded in the compiled program and cannot be mutated post-compilation. Na\"ively, this means every weight update during training requires full recompilation. However, as we show in Section~\ref{sec:delta}, compiled programs can be surgically updated by exploiting the unload/reload interface of \texttt{\_ANEModel}, bypassing the compiler entirely.

\subsection{The Private API Model}

Table~\ref{tab:private_api} lists the key private classes used by \textsc{Orion}. These are loaded at runtime via \texttt{dlopen()} and \texttt{objc\_getClass()} from \texttt{/System/Library/PrivateFrameworks/AppleNeuralEngine.framework}.

\begin{table}[h]
\centering
\caption{Private ANE API surface used by \textsc{Orion}.}
\label{tab:private_api}
\begin{tabular}{ll}
\toprule
\textbf{Class} & \textbf{Role} \\
\midrule
\texttt{\_ANEClient} & Singleton connection to ANE daemon \\
\texttt{\_ANECompiler} & MIL $\to$ E5 microcode compilation \\
\texttt{\_ANEInMemoryModel} & In-memory model (no filesystem) \\
\texttt{\_ANEInMemoryModelDescriptor} & Accepts MIL + weight blobs \\
\texttt{\_ANEModel} & Holds compiled program handle \\
\texttt{\_ANERequest} & Evaluation request specification \\
\texttt{\_ANEIOSurfaceObject} & IOSurface wrapper for tensor I/O \\
\bottomrule
\end{tabular}
\end{table}

\subsection{Prior Work}

Three projects form the foundation for \textsc{Orion}:

\textbf{maderix}~\citep{maderix2024ane,maderix2024ane_part1,maderix2024ane_part2} made the foundational contribution to direct ANE programming. The project reverse-engineered the private API calling sequence (compile $\to$ load $\to$ evaluate), demonstrated raw ANE dispatch from C, and produced the first empirical hardware characterization of the M4 ANE. Key findings include: (1) debunking Apple's 38 TOPS specification by showing that INT8 is dequantized to fp16 before computation, yielding $\sim$19 TFLOPS actual throughput; (2) identifying the 32 MB SRAM performance cliff (30\% throughput drop when exceeded); (3) measuring XPC+IOKit dispatch overhead at $\sim$0.095 ms; (4) showing that 1$\times$1 convolutions deliver 3$\times$ better throughput than equivalent \texttt{matmul} operations; (5) discovering that deep operation graphs (16--64 ops) achieve 94\% ANE utilization versus $\sim$30\% for single operations; and (6) identifying the $\sim$119 compilation-per-process limit. These findings established both the feasibility and the performance envelope of direct ANE programming.

\textbf{ANEgpt}~\citep{anegpt2024} extended the maderix APIs to transformer training, implementing forward and backward passes for a 110M-parameter model on ANE. However, ANEgpt suffers from NaN divergence after the first training step on resume, uses Python for orchestration, generates MIL through string concatenation, and lacks a compiler or optimization pipeline.

\textbf{hollance/neural-engine}~\citep{hollance2022ane} documented ANE characteristics through CoreML-level experiments, establishing the preference for \texttt{[B, C, 1, S]} tensor layouts and providing early guidance on which operations are ANE-friendly.

\textsc{Orion} builds on all three: it uses the API calling sequence and hardware characterization from maderix, the training kernel structure from ANEgpt, and the layout insights from hollance. \textsc{Orion}'s contributions beyond this foundation are: (a) a compiler with graph IR, optimization passes, and verified code generation; (b) stable multi-step training with checkpoint resume (solving three NaN-inducing bugs); (c) 14 newly discovered MIL IR and memory constraints; and (d) a complete Objective-C runtime (Python is used only for one-time weight conversion) with inference, training, benchmarking, and native tokenizers.

\section{ANE Characterization}
\label{sec:characterization}

Table~\ref{tab:constraints} presents a consolidated catalog of 20 ANE programming constraints. Six of these (\#5, 7, 15, 17, and partially \#6) were first documented by maderix~\citep{maderix2024ane_part1,maderix2024ane_part2} or hollance~\citep{hollance2022ane} through hardware-level benchmarking and API exploration. The remaining 14 constraints were discovered during \textsc{Orion} development through 161 engineering tasks spanning 18 sessions, primarily involving MIL IR compilation failures, evaluation errors, and silent numerical corruption encountered while building the compiler, training loop, delta compilation, and LoRA adapter system.

\begin{table*}[t]
\centering
\caption{ANE constraint catalog. Source: P = prior work~\citep{maderix2024ane_part1,maderix2024ane_part2,hollance2022ane}, O = discovered during \textsc{Orion} development, $^*$confirmed by maderix/ANEgpt codebases. Prior-work constraints were independently confirmed on M4 Max.}
\label{tab:constraints}
\footnotesize
\setlength{\tabcolsep}{4pt}
\begin{tabular}{clp{2.6cm}p{3.2cm}c}
\toprule
\textbf{\#} & \textbf{Constraint} & \textbf{Symptom} & \textbf{Workaround} & \textbf{Src} \\
\midrule
1 & \texttt{concat} MIL op rejected by ANE compiler & Compile failure & Split into separate programs & O \\
2 & Multi-output buffers must have uniform sizes & \texttt{0x1d} at eval & Pad outputs to max size & O \\
3 & Multi-output surfaces ordered alphabetically & Silent wrong data & Name outputs in sorted order & O \\
4 & Minimum $\sim$49\,KB IOSurface for eval & \texttt{0x1d} at eval & Pad seq dim $\geq$ 16 & O \\
5 & $\sim$119 compilations per process limit & Silent fail / crash & \texttt{exec()} restart & P \\
6 & SDPA causal masks silently ignored & Wrong attention & Manual causal masking & P$^*$ \\
7 & Weights baked at compile time & Stale weights & Recompile after update & P \\
8 & BLOBFILE offset is \texttt{uint64(64)}, not 128 & Garbage weights & Correct offset in MIL ref & O \\
9 & MIL text must be \texttt{NSData*}, not \texttt{NSString*} & Immediate crash & Encode to UTF-8 data & O \\
10 & \texttt{gelu} is not a valid MIL activation & Compile failure & Tanh approximation & O \\
11 & Weight dict must be \texttt{@\{\}}, not \texttt{nil} & Immediate crash & Pass empty dictionary & O \\
12 & \texttt{matmul} transpose flags need named consts & MIL rejection & Emit \texttt{const} nodes & O \\
13 & \texttt{conv} does not support \texttt{bias=} param & MIL rejection & Separate add operation & O \\
14 & Output vars must ref live (post-opt) nodes & Invalid program & Update refs after DCE & O \\
15 & \texttt{exec()} restart overhead $\sim$50\,ms & Latency cost & Batch steps per process & P \\
16 & 32K-channel convolutions rejected & Compile failure & CPU fallback or chunking & O \\
17 & Conv 1$\times$1 is 3$\times$ faster than \texttt{matmul} & Performance gap & Prefer conv formulation & P \\
18 & Multi-input surfaces must have uniform alloc sizes & \texttt{0x1d} at eval & Allocate all inputs at max size & O \\
19 & Multi-input surfaces ordered alphabetically & Silent wrong data & Name inputs in sorted order & O \\
20 & ANE reads flat buffer as packed \texttt{[1,C,1,S]} & Silent wrong data & Write packed data at buffer start & O \\
\bottomrule
\end{tabular}
\end{table*}

We organize these constraints into four categories:

\paragraph{MIL IR Restrictions (\#1, 6, 10, 12, 13, 16).} The ANE compiler accepts a subset of MIL operations. Several operations that are valid in CoreML's MIL specification are silently rejected or produce incorrect results on the ANE. Most critically, the \texttt{concat} operation (\#1) causes immediate compilation failure, requiring all multi-tensor operations to be decomposed into separate programs. The \texttt{gelu} activation (\#10) must be replaced with its tanh approximation: $\text{GELU}(x) \approx 0.5x(1 + \tanh[\sqrt{2/\pi}(x + 0.044715x^3)])$.

\paragraph{Memory and I/O Constraints (\#2, 3, 4, 8, 9, 11, 18, 19, 20).} The ANE has strict requirements on tensor memory layout. Multi-output programs require all output buffers to have identical byte sizes (\#2), with outputs ordered alphabetically by their MIL variable names (\#3). Symmetrically, multi-input programs require all input IOSurfaces to have the same allocation size (\#18), with inputs also ordered alphabetically by MIL parameter name (\#19). When input surfaces are over-allocated (padded to uniform size), the ANE reads the flat buffer as packed \texttt{[1,C,1,S]} data starting from byte 0, ignoring the surface's nominal dimensions (\#20). There is a minimum IOSurface size of approximately 49 KB (\#4), meaning single-token tensors with shape \texttt{[1, 768, 1, 1]} (3,072 bytes in fp16) must be padded to at least \texttt{[1, 768, 1, 16]} (24,576 bytes). The BLOBFILE weight format uses an offset of 64 bytes from the chunk header (\#8), not from the file start --- an undocumented detail that causes silent weight corruption if incorrect. Constraints \#18--20 were discovered during LoRA implementation (Section~\ref{sec:lora}), where adapter matrices of different shapes must be passed as multiple IOSurface inputs to a single program.

\paragraph{Compilation Limits (\#5, 7, 14, 15).} The ANE compiler maintains internal state that limits each process to approximately 119 compilations before subsequent compilations silently fail (\#5). Since weights are baked at compile time (\#7), every training step requires recompilation of weight-bearing kernels. \textsc{Orion} v1.0 addressed this with an \texttt{exec()} restart strategy: after each training step, the process re-executes itself with updated checkpoint state, resetting the compilation counter at a cost of $\sim$50 ms (\#15). \textsc{Orion} v2.0 eliminates this constraint entirely via delta compilation (Section~\ref{sec:delta}), which bypasses \texttt{ANECCompile()} by reloading existing program objects with updated weight files.

\paragraph{Performance Characteristics (\#16, 17).} The ANE's convolution engine delivers $\sim$3$\times$ better throughput for 1$\times$1 convolutions compared to equivalent \texttt{matmul} operations (\#17), first measured by maderix~\citep{maderix2024ane_part2} and also noted by~\citet{hollance2022ane}. However, convolutions with very large channel counts (e.g., 32,000 for vocabulary projection) are rejected (\#16), a new finding that requires CPU fallback for classifier layers.

\section{System Design}
\label{sec:system}

\begin{figure}[h]
\centering
\begin{tikzpicture}[
  layer/.style={draw, rounded corners=3pt, minimum width=10cm, minimum height=0.9cm, text centered, font=\small},
  arrow/.style={-{Stealth[length=2.5mm]}, thick},
  node distance=0.35cm
]

\node[layer, fill=red!8] (ane) {Apple Neural Engine (private APIs: \texttt{\_ANEClient}, \texttt{\_ANECompiler}, MIL IR)};
\node[layer, fill=orange!10, above=of ane] (runtime) {Core Runtime (compile, eval, IOSurface I/O, program cache)};
\node[layer, fill=yellow!10, above=of runtime] (compiler) {Compiler (Graph IR $\to$ 5 optimization passes $\to$ MIL codegen)};
\node[layer, fill=green!8, above=of compiler] (model) {Model Layer (configs, tokenizers, weight loading)};
\node[layer, fill=blue!8, above=of model] (cli) {CLI Applications (infer, train, bench)};

\draw[arrow] (cli) -- (model);
\draw[arrow] (model) -- (compiler);
\draw[arrow] (compiler) -- (runtime);
\draw[arrow] (runtime) -- (ane);

\node[right=0.3cm of cli, font=\scriptsize\itshape, text=gray] {User-facing};
\node[right=0.3cm of compiler, font=\scriptsize\itshape, text=gray] {27 ops, 13 frontends};
\node[right=0.3cm of runtime, font=\scriptsize\itshape, text=gray] {Zero-copy, delta compilation};
\node[right=0.3cm of ane, font=\scriptsize\itshape, text=gray] {E5 microcode};

\end{tikzpicture}
\caption{Orion architecture stack. Each layer communicates only with its immediate neighbors. The compiler and runtime together abstract away ANE constraints from the model layer.}
\label{fig:architecture}
\end{figure}
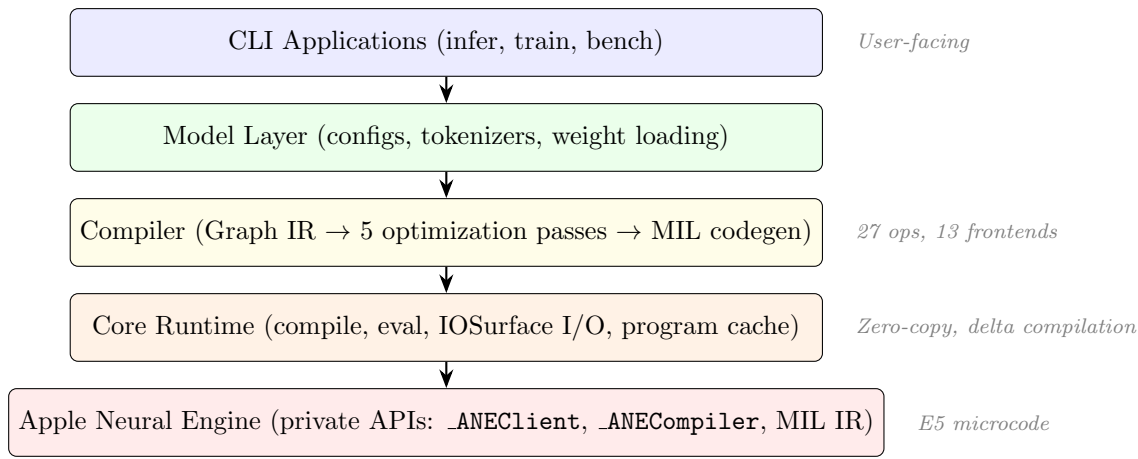

\textsc{Orion} is structured as five layers, shown in Figure~\ref{fig:architecture}.

\subsection{Compiler Pipeline}

The \textsc{Orion} compiler transforms a high-level graph IR into ANE-executable MIL programs. The graph IR supports 27 operations (Table~\ref{tab:graph_ops}) and maintains explicit tensor shape information for all edges.

\begin{table}[h]
\centering
\caption{Graph IR operation categories (27 total).}
\label{tab:graph_ops}
\small
\begin{tabular}{ll}
\toprule
\textbf{Category} & \textbf{Operations} \\
\midrule
Data & \texttt{input}, \texttt{const}, \texttt{identity} \\
Linear & \texttt{conv1x1}, \texttt{matmul} \\
Elementwise & \texttt{add}, \texttt{sub}, \texttt{mul}, \texttt{neg} \\
Activation & \texttt{relu}, \texttt{tanh}, \texttt{sigmoid} \\
Math & \texttt{exp}, \texttt{pow}, \texttt{sqrt}, \texttt{rsqrt} \\
Reduction & \texttt{reduce\_sum}, \texttt{reduce\_mean}, \texttt{reduce\_max} \\
Shape & \texttt{reshape}, \texttt{transpose}, \texttt{split}, \texttt{pad}, \texttt{slice} \\
Other & \texttt{cast}, \texttt{softmax}, \texttt{concat\_banned} \\
\bottomrule
\end{tabular}
\end{table}

The optimization pipeline runs five passes in a fixpoint loop (maximum 20 iterations):

\begin{enumerate}[leftmargin=*,itemsep=2pt]
  \item \textbf{Dead Code Elimination (DCE).} Marks nodes reachable from outputs via backward walk; removes unreachable nodes.
  \item \textbf{Identity Elimination.} Removes no-op casts (same type), reshapes (same shape), and identity transpositions.
  \item \textbf{Cast Fusion.} Eliminates round-trip casts (e.g., fp16$\to$fp32$\to$fp16) that arise from mixed-precision patterns.
  \item \textbf{SRAM Annotation.} Estimates working-set size against the 32 MB on-chip SRAM budget; emits warnings when exceeded (performance degrades $\sim$30\%).
  \item \textbf{ANE Constraint Validation.} Checks for banned operations (\texttt{concat}), minimum tensor sizes, weight dictionary requirements, and output variable liveness.
\end{enumerate}

After optimization, MIL codegen emits text-format MIL programs with BLOBFILE weight references. The compiler includes 13 verified frontends covering GPT-2 inference (prefill attention, prefill FFN, decode projection, decode FFN, final LayerNorm) and Stories110M training (forward attention, forward FFN, FFN backward, SDPA backward parts 1 and 2, QKV backward, classifier forward, vocabulary softmax). All 13 frontends have been verified structurally equivalent to hand-written MIL via an automated diff tool.

\subsection{Runtime}

The runtime manages the lifecycle of ANE programs: compilation, caching, evaluation, and I/O. Key design decisions include:

\textbf{Program cache.} Compiled programs are cached with composite keys (model name, layer index, sequence length, weight version). Cache hits skip the $\sim$11 ms compilation overhead per program.

\textbf{IOSurface I/O.} All tensor data resides in IOSurface-backed memory, enabling zero-copy sharing between the CPU address space and the ANE. The runtime handles the transpose between CPU-native \texttt{[seq, d\_model]} and ANE-native \texttt{[1, d\_model, 1, seq]} layouts.

\textbf{Delta compilation.} Rather than recompiling programs after each weight update, the runtime uses a surgical reload approach: unload the existing program from the ANE, update weight files (BLOBFILEs) on disk, and reload. This bypasses \texttt{ANECCompile()} entirely and eliminates the $\sim$119 compilation limit (Section~\ref{sec:delta}). The original \texttt{exec()} restart strategy (v1.0) has been superseded.

\subsection{CPU/ANE Division of Labor}

Not all operations can or should run on the ANE. Table~\ref{tab:division} shows the division of labor in \textsc{Orion}.

\begin{table}[h]
\centering
\caption{CPU/ANE work division. Operations are assigned based on ANE hardware constraints and performance characteristics.}
\label{tab:division}
\small
\begin{tabular}{lll}
\toprule
\textbf{Operation} & \textbf{Device} & \textbf{Reason} \\
\midrule
Transformer fwd/bwd (dx) & ANE & Compute-bound convolutions \\
Token sampling & CPU & Sequential, branching logic \\
Adam optimizer & CPU & Weights immutable on ANE \\
$\nabla W$ accumulation & CPU & cblas\_sgemm via GCD \\
NLL loss + gradient & CPU & \texttt{gather} not in MIL \\
Classifier backward & CPU & 32K channels rejected \\
Embedding lookup & CPU & Table indexing \\
\bottomrule
\end{tabular}
\end{table}

\subsection{Inference Pipeline}

For GPT-2 124M inference, \textsc{Orion} implements bucketed prefill followed by autoregressive decode:

\begin{enumerate}[leftmargin=*,itemsep=1pt]
  \item \textbf{Prefill.} The prompt is tokenized, embedded on CPU, then processed through all 12 transformer layers on the ANE using prefill programs with sequence-length buckets (32, 64, 128, 256, 512, 1024). The KV cache is populated.
  \item \textbf{Decode.} Each subsequent token is processed through the full model on ANE with a minimum sequence dimension of 16 (to satisfy constraint \#4). The KV cache is updated incrementally.
  \item \textbf{Sampling.} Logits are returned to CPU for temperature/top-$p$ sampling.
\end{enumerate}

First-call latency includes ANE compilation ($\sim$1015 ms for 24 programs); subsequent calls use cached programs.

\subsection{Training Pipeline}

For Stories110M training on TinyStories~\citep{eldan2023tinystories}:

\begin{enumerate}[leftmargin=*,itemsep=1pt]
  \item At startup, 72 ANE programs are compiled once (60 weight-bearing + 12 static SDPA backward kernels, 6 per layer). This is the only compilation in the entire training run.
  \item Forward pass: ANE executes \texttt{fwdAttn} (RMSNorm $\to$ QKV $\to$ SDPA $\to$ $W_o$) and \texttt{fwdFFN} (RMSNorm $\to$ SwiGLU) per layer.
  \item Loss: CPU computes NLL loss and gradient.
  \item Backward pass (dx): ANE executes \texttt{ffnBwd}, \texttt{sdpaBwd1}, \texttt{sdpaBwd2}, \texttt{qkvBwd} per layer.
  \item Weight gradients: CPU computes $\nabla W$ via \texttt{cblas\_sgemm} with GCD parallelism.
  \item Adam update on CPU; delta reload of 60 weight-bearing programs ($\sim$494\,ms total). No recompilation, no process restart.
\end{enumerate}

\section{Delta Compilation}
\label{sec:delta}

The ANE's compile-then-dispatch model creates a fundamental tension with gradient descent: every weight update requires new weights to be baked into compiled programs. In \textsc{Orion} v1.0, this meant full recompilation of 60 weight-bearing kernels per training step ($\sim$4,200\,ms), consuming 83.9\% of wall time. The $\sim$119 compilation-per-process limit (\#5 in Table~\ref{tab:constraints}) further required \texttt{exec()} restart after every step.

\subsection{Key Insight}

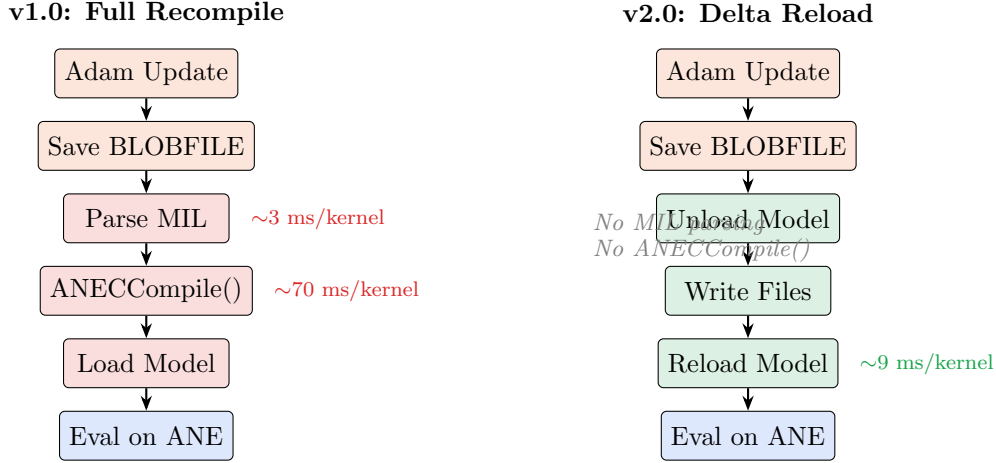
\begin{figure}[h]
\centering
\begin{tikzpicture}[
  box/.style={draw, rounded corners=2pt, minimum width=2.2cm, minimum height=0.65cm, font=\small, text centered},
  arrow/.style={-{Stealth[length=2mm]}, thick},
  node distance=0.3cm and 0.6cm,
]

\node[font=\small\bfseries] at (-2.8, 3.2) {v1.0: Full Recompile};
\node[box, fill=cpuorange!15] (adam1) at (-2.8, 2.4) {Adam Update};
\node[box, fill=cpuorange!15, below=of adam1] (save1) {Save BLOBFILE};
\node[box, fill=nanred!15, below=of save1] (mil1) {Parse MIL};
\node[box, fill=nanred!15, below=of mil1] (compile1) {ANECCompile()};
\node[box, fill=nanred!15, below=of compile1] (load1) {Load Model};
\node[box, fill=aneblue!15, below=of load1] (eval1) {Eval on ANE};

\draw[arrow] (adam1) -- (save1);
\draw[arrow] (save1) -- (mil1);
\draw[arrow] (mil1) -- (compile1);
\draw[arrow] (compile1) -- (load1);
\draw[arrow] (load1) -- (eval1);

\node[right=0.15cm of compile1, font=\scriptsize, text=nanred] {$\sim$70 ms/kernel};
\node[right=0.15cm of mil1, font=\scriptsize, text=nanred] {$\sim$3 ms/kernel};

\node[font=\small\bfseries] at (5.2, 3.2) {v2.0: Delta Reload};
\node[box, fill=cpuorange!15] (adam2) at (5.2, 2.4) {Adam Update};
\node[box, fill=cpuorange!15, below=of adam2] (save2) {Save BLOBFILE};
\node[box, fill=fixgreen!15, below=of save2] (unload2) {Unload Model};
\node[box, fill=fixgreen!15, below=of unload2] (write2) {Write Files};
\node[box, fill=fixgreen!15, below=of write2] (reload2) {Reload Model};
\node[box, fill=aneblue!15, below=of reload2] (eval2) {Eval on ANE};

\draw[arrow] (adam2) -- (save2);
\draw[arrow] (save2) -- (unload2);
\draw[arrow] (unload2) -- (write2);
\draw[arrow] (write2) -- (reload2);
\draw[arrow] (reload2) -- (eval2);

\node[right=0.15cm of reload2, font=\scriptsize, text=fixgreen] {$\sim$9 ms/kernel};

\node[font=\footnotesize, text=gray, anchor=west] at (3.0, 0.4) {\emph{No MIL parsing}};
\node[font=\footnotesize, text=gray, anchor=west] at (3.0, 0.05) {\emph{No ANECCompile()}};

\end{tikzpicture}
\caption{Weight update paths. v1.0 (left) creates new model descriptors and invokes the ANE compiler for every weight update ($\sim$70\,ms/kernel). v2.0 (right) reuses existing model objects: unload, write new weight files, reload ($\sim$9\,ms/kernel). The compiler is bypassed entirely.}
\label{fig:delta_arch}
\end{figure}

Figure~\ref{fig:delta_arch} contrasts the v1.0 and v2.0 weight update paths.

Compiled ANE programs are managed by \texttt{\_ANEModel} objects that expose \texttt{unloadWithQoS:} and \texttt{loadWithQoS:} methods. When a model is unloaded, its backing weight files (BLOBFILEs) on disk can be modified. Reloading the model picks up the new weights \emph{without invoking \texttt{ANECCompile()}} --- the E5 microcode and MIL text are unchanged; only the weight data is refreshed. Crucially, when the MIL text and weight dictionary keys are identical, the ANE assigns the same \texttt{hexStringIdentifier} (a composite of three SHA-256 hashes), so the program's internal identity is preserved across reloads.

\subsection{Implementation}

Algorithm~\ref{alg:delta} shows the delta compilation procedure. For each of the 60 weight-bearing kernels:

\begin{algorithm}[h]
\caption{Delta compilation (weight reload)}
\label{alg:delta}
\begin{algorithmic}[1]
\Require Compiled program $P$ with model handle $M$, new weight dict $W'$
\State $M$.\texttt{unloadWithQoS(21)} \Comment{Remove from ANE}
\For{each weight file path $p$ in $W'$}
  \State Write $W'[p]$ to disk at $M$.\texttt{tmpDir}/$p$ \Comment{Update BLOBFILE}
\EndFor
\State $M$.\texttt{loadWithQoS(21)} \Comment{Reload with new weights}
\end{algorithmic}
\end{algorithm}

This replaces the full compilation path: no \texttt{\_ANEInMemoryModelDescriptor} creation, no MIL parsing, no \texttt{ANECCompile()} invocation. The implementation (\texttt{orion\_program\_reload\_weights} in \texttt{core/ane\_runtime.m}) handles ownership transfer of the temporary directory between old and new program states.

\subsection{Results}

\begin{figure}[h]
\centering
\begin{tikzpicture}
\begin{axis}[
  width=0.95\columnwidth,
  height=6cm,
  ybar stacked,
  xlabel={Training Configuration},
  ylabel={Time per Step (ms)},
  symbolic x coords={v1.0 Full Recompile, v2.0 Delta Reload},
  xtick=data,
  ymin=0, ymax=6000,
  bar width=40pt,
  enlarge x limits=0.4,
  legend style={at={(0.5,0.97)}, anchor=north, legend columns=2, font=\small},
  grid=major,
  grid style={gray!20},
  nodes near coords,
  nodes near coords style={font=\scriptsize},
  every node near coord/.append style={anchor=south},
]

\addplot[fill=aneblue!70, draw=aneblue] coordinates {
  (v1.0 Full Recompile, 908)
  (v2.0 Delta Reload, 849)
};

\addplot[fill=cpuorange!70, draw=cpuorange] coordinates {
  (v1.0 Full Recompile, 4200)
  (v2.0 Delta Reload, 494)
};

\legend{Compute, Recompile/Reload}
\end{axis}
\end{tikzpicture}
\caption{Training step time breakdown. v1.0 spends 83.9\% of each step on full ANE recompilation ($\sim$4,200\,ms for 60 kernels). v2.0's delta reload reduces this to 494\,ms by bypassing \texttt{ANECCompile()} entirely, yielding a 3.8$\times$ total speedup.}
\label{fig:delta_comparison}
\end{figure}
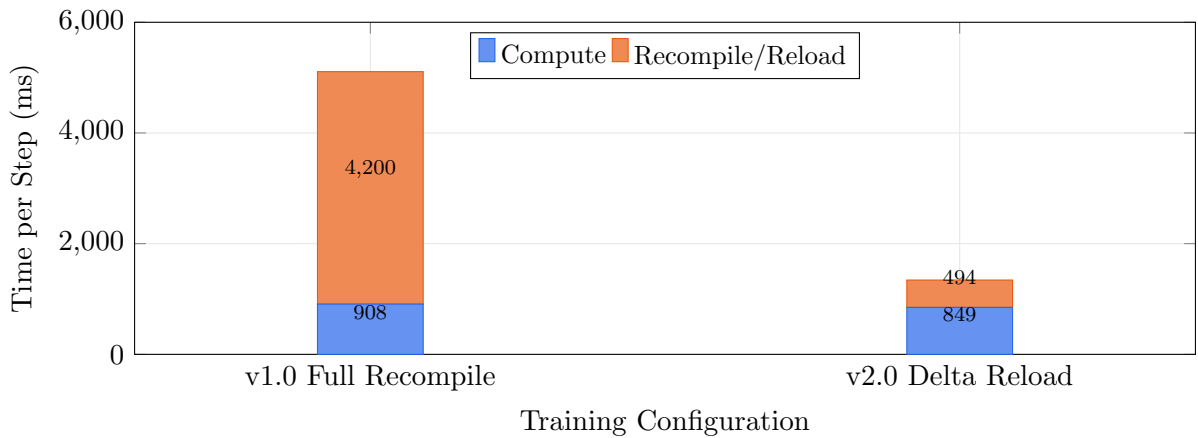

Figure~\ref{fig:delta_comparison} and Table~\ref{tab:delta} compare v1.0 (full recompile) and v2.0 (delta reload) training performance.

\begin{table}[h]
\centering
\caption{Training step time breakdown: v1.0 (full recompile) vs v2.0 (delta reload). Stories110M on M4 Max, lr=$3\!\times\!10^{-4}$, grad\_accum=4.}
\label{tab:delta}
\begin{tabular}{lrrr}
\toprule
\textbf{Metric} & \textbf{v1.0} & \textbf{v2.0} & \textbf{Speedup} \\
\midrule
Train time (compute) & 908 ms & 849 ms & $\sim$1$\times$ \\
Recompile / reload time & 4,200 ms & 494 ms & 8.5$\times$ \\
Total step time & 5,108 ms & 1,345 ms & 3.8$\times$ \\
Recompile \% of step & 83.9\% & 36.8\% & $-$47.1 pp \\
1000-step wall time & $\sim$85 min & 22.4 min & 3.8$\times$ \\
Process model & 1 step/process & Single process & --- \\
Compiles during training & 72/step & 0 & Eliminated \\
\bottomrule
\end{tabular}
\end{table}

The 8.5$\times$ recompile speedup comes from avoiding three expensive operations in the full compilation path: (1)~creating new \texttt{\_ANEInMemoryModelDescriptor} objects ($\sim$3\,ms/kernel for MIL parsing), (2)~invoking \texttt{ANECCompile()} ($\sim$30--80\,ms/kernel), and (3)~loading a new model identity ($\sim$30\,ms/kernel). Delta reload replaces all three with a single unload--write--reload cycle ($\sim$8\,ms/kernel).

A critical engineering detail: when the new and old programs share the same \texttt{hexStringIdentifier} (which they do, since the MIL text is unchanged), they share the same temporary directory on disk. The old program's ownership of this directory must be transferred before release, or the \texttt{orion\_release\_program} destructor will delete the shared directory, causing the new program to fail on its next reload.

\section{LoRA Adapter-as-Input}
\label{sec:lora}

Since the ANE bakes weights at compile time, adapting a model to new tasks traditionally requires full recompilation. We implement LoRA~\citep{hu2022lora} with a key architectural decision: adapter matrices $A$ and $B$ are passed as \emph{IOSurface inputs} rather than baked weights. This enables hot-swap of adapters without any recompilation.

\subsection{Architecture}

For a linear layer $Y = XW_{\text{base}}$, the LoRA-fused computation is:
\begin{equation}
Y = XW_{\text{base}} + \alpha \cdot (XA)B
\end{equation}
where $W_{\text{base}} \in \mathbb{R}^{d \times d}$ is baked as a BLOBFILE weight, and $A \in \mathbb{R}^{d \times r}$, $B \in \mathbb{R}^{r \times d}$ (rank $r \ll d$) are IOSurface inputs. The base weights remain compiled into the program; only the low-rank adapters are passed at evaluation time.

The \textsc{Orion} compiler includes two LoRA frontends:
\begin{itemize}[leftmargin=*,itemsep=1pt]
  \item \texttt{orion\_frontend\_lora\_linear}: Single linear layer with LoRA fusion.
  \item \texttt{orion\_frontend\_lora\_attention}: Full attention block with LoRA on Q, K, V, and O projections (8 adapter matrices as IOSurface inputs).
\end{itemize}

\subsection{IOSurface Input Constraints}

Implementing LoRA revealed three new ANE constraints (\#18--20 in Table~\ref{tab:constraints}):

\begin{enumerate}[leftmargin=*,itemsep=1pt]
  \item \textbf{Uniform input allocation} (\#18). All IOSurface inputs to a single program must have the same byte allocation size, even if the underlying tensors have different shapes. For LoRA attention with 8 adapter matrices of varying dimensions, all inputs are allocated at the maximum size.
  \item \textbf{Alphabetical input ordering} (\#19). Input IOSurfaces are bound to MIL parameters in alphabetical order by parameter name, not by the order in which they appear in the function signature.
  \item \textbf{Packed flat reads} (\#20). When an input surface is over-allocated (padded to uniform size), the ANE reads the flat buffer from byte 0 as packed \texttt{[1,C,1,S]} data, ignoring the surface's nominal dimensions. Adapter data must be written starting at the buffer's beginning.
\end{enumerate}

\subsection{Hot-Swap}

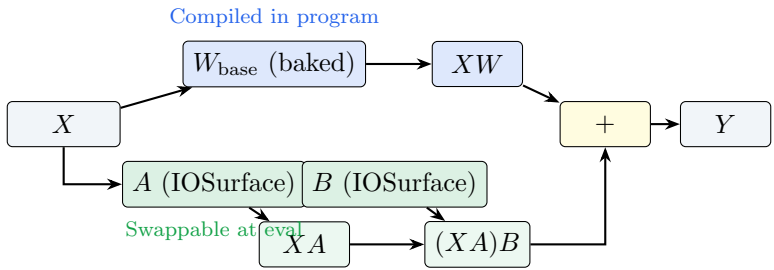
\begin{figure}[h]
\centering
\begin{tikzpicture}[
  box/.style={draw, rounded corners=2pt, minimum height=0.6cm, font=\small, text centered},
  iobox/.style={box, fill=fixgreen!15, minimum width=1.8cm},
  baked/.style={box, fill=aneblue!15, minimum width=2.2cm},
  arrow/.style={-{Stealth[length=2mm]}, thick},
]

\node[box, fill=layergray, minimum width=1.5cm] (x) at (0, 0) {$X$};

\node[baked] (wbase) at (2.8, 0.8) {$W_{\text{base}}$ (baked)};
\node[box, fill=aneblue!15, minimum width=1.2cm] (xw) at (5.5, 0.8) {$XW$};

\node[iobox] (a) at (2.0, -0.8) {$A$ (IOSurface)};
\node[iobox] (b) at (4.4, -0.8) {$B$ (IOSurface)};
\node[box, fill=fixgreen!8, minimum width=1.2cm] (xa) at (3.2, -1.6) {$XA$};
\node[box, fill=fixgreen!8, minimum width=1.2cm] (xab) at (5.5, -1.6) {$(XA)B$};

\node[box, fill=yellow!15, minimum width=1.2cm] (sum) at (7.2, 0) {$+$};
\node[box, fill=layergray, minimum width=1.2cm] (y) at (8.8, 0) {$Y$};

\draw[arrow] (x) -- (wbase);
\draw[arrow] (wbase) -- (xw);
\draw[arrow] (x) |- (a);
\draw[arrow] (a) -- (xa);
\draw[arrow] (b) -- (xab);
\draw[arrow] (xa) -- (xab);
\draw[arrow] (xw) -- (sum);
\draw[arrow] (xab) -| (sum);
\draw[arrow] (sum) -- (y);

\node[font=\scriptsize, text=aneblue, above=0.05cm of wbase] {Compiled in program};
\node[font=\scriptsize, text=fixgreen, below=0.05cm of a] {Swappable at eval};

\end{tikzpicture}
\caption{LoRA-fused linear layer. Base weights $W_{\text{base}}$ are baked into the compiled ANE program (blue). Adapter matrices $A$, $B$ are passed as IOSurface inputs (green) and can be swapped without recompilation. $Y = XW_{\text{base}} + \alpha(XA)B$.}
\label{fig:lora_arch}
\end{figure}

Once a base program is compiled with LoRA-fused frontends, swapping to a different adapter requires only changing the IOSurface input data --- zero recompilation, zero program cache invalidation. Figure~\ref{fig:lora_arch} illustrates the data flow.

The \texttt{OrionLoRAAdapter} struct holds pre-allocated IOSurface tensors for all adapter matrices, loaded from BLOBFILE-format files via \texttt{orion\_lora\_load()}.

\section{Numerical Stability}
\label{sec:numerical}

Achieving stable training on the ANE required solving three interacting bugs that caused 100\% NaN divergence after the first training step in the upstream ANEgpt system.

\subsection{Bug 1: Stale Programs on Resume}

\textbf{Root cause.} ANE programs were compiled \emph{before} checkpoint weights were loaded. The forward pass used stale (pre-checkpoint) weights, while the backward pass expected gradients consistent with the new weights. This created a weight mismatch that diverged within one step.

\textbf{Fix: Deferred compilation.} Programs are now compiled \emph{after} checkpoint loading, ensuring the weights baked into each program match the current optimizer state. Each process compiles exactly once with the correct weights.

\subsection{Bug 2: fp16 Overflow Cascade}

\textbf{Root cause.} The ANE operates natively in fp16 ($\pm$65,504 dynamic range). Large intermediate activations overflowed to $\pm\infty$, which propagated through softmax and cross-entropy to produce NaN loss values.

\textbf{Fix: Activation clamping.} Before softmax and layer normalization, activations are clamped to $[-65504, +65504]$:
\begin{equation}
\hat{x}_i = \text{clamp}(x_i, -65504, +65504)
\end{equation}
This prevents overflow without affecting well-behaved activations (which are orders of magnitude smaller than the fp16 limit).

\subsection{Bug 3: Corrupted BLOBFILE Weights}

\textbf{Root cause.} The BLOBFILE writer produced corrupted weight data when checkpoint tensor layouts did not match the expected MIL weight dictionary format. This caused silent numerical corruption --- weights loaded without error but contained garbage values.

\textbf{Fix: Gradient sanitization.} Before writing to BLOBFILE, all gradient values are sanitized: NaN $\to$ 0, $\pm\infty \to \pm$65504. Additionally, a validation pass detects corrupted weights early by checking for NaN/Inf values after BLOBFILE load.

\subsection{Combined Effect}

\begin{figure}[h]
\centering
\begin{tikzpicture}
\begin{axis}[
  width=0.85\columnwidth,
  height=5.5cm,
  xlabel={Training Step},
  ylabel={Loss},
  xmin=0.5, xmax=5.5,
  ymin=13.5, ymax=14.5,
  xtick={1,2,3,4,5},
  legend pos=north east,
  legend style={font=\small},
  grid=major,
  grid style={gray!20},
]

\addplot[color=nanred, mark=*, thick, mark size=2.5pt, dashed] coordinates {
  (1, 13.98)
};
\node[font=\footnotesize, color=nanred, fill=white, draw=nanred, rounded corners=2pt, inner sep=2pt] at (axis cs:2.6,14.3) {ANEgpt: NaN at step 2};
\draw[-{Stealth[length=2mm]}, nanred, thick] (axis cs:2.05, 14.2) -- (axis cs:2.05, 14.45);
\draw[nanred, thick, dashed] (axis cs:1, 13.98) -- (axis cs:2.05, 14.2);

\addplot[color=fixgreen, mark=square*, thick, mark size=2.5pt] coordinates {
  (1, 13.98)
  (2, 13.97)
  (3, 13.95)
  (4, 13.94)
  (5, 13.92)
};

\legend{Before fix (ANEgpt), After fix (Orion)}
\end{axis}
\end{tikzpicture}
\caption{Training loss before and after the three-bug NaN fix. ANEgpt diverges to NaN at step 2 with 100\% reproducibility (red, dashed arrow indicates divergence to $\infty$). \textsc{Orion} achieves stable, monotonically decreasing loss across 5 steps with checkpoint resume (green).}
\label{fig:nanfix}
\end{figure}
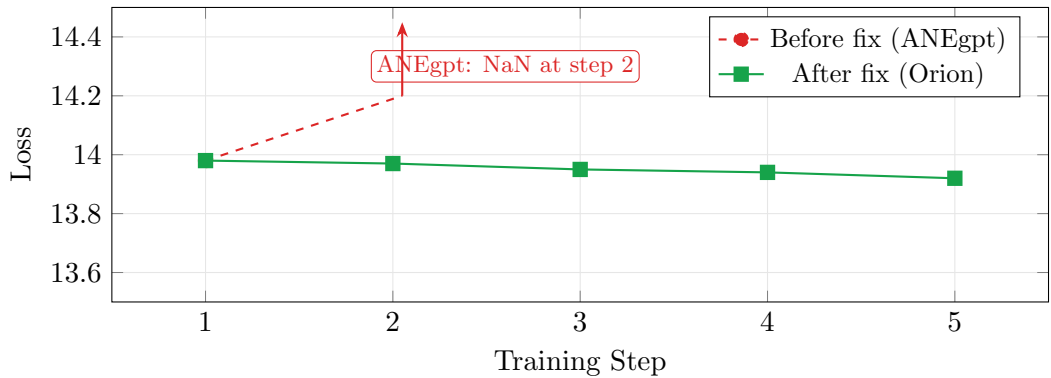

Figure~\ref{fig:nanfix} shows the training loss before and after these fixes. The upstream system (ANEgpt) diverges to NaN at step 2 with 100\% reproducibility. After the three-bug fix, \textsc{Orion} achieves stable training for 1,000 steps with zero NaN occurrences, verified across a 5-chain stress test, a v1.0 1,000-step run (loss: 12.3$\to$6.2, $\sim$85 min), and a v2.0 1,000-step run with delta compilation (loss: 12.3$\to$8.9, 22 min) (Section~\ref{sec:evaluation}).

\subsection{Stability Validation}

To move beyond a single 5-step anecdote, we designed a structured stability stress test that exercises the full resume pipeline. We run 5 independent resume chains, each starting from the same pretrained weights, with each step executing in a fresh process (via \texttt{exec()} restart) to exercise the full checkpoint--load--compile--train--save cycle.

\begin{table}[h]
\centering
\caption{Training stability stress test results (Stories110M, lr=$10^{-5}$, grad\_accum=4, M4 Max). Each chain is an independent 5-step resume sequence.}
\label{tab:stability}
\small
\begin{tabular}{lr}
\toprule
\textbf{Metric} & \textbf{Value} \\
\midrule
Resume chains tested & 5 \\
Steps per chain & 5 (each in a fresh process) \\
Total training steps & 25 \\
NaN / Inf occurrences & 0 / 25 \\
Loss monotonically decreasing & 5 / 5 chains \\
Step 1 loss (mean $\pm$ std across chains) & $13.975 \pm 0.003$ \\
Step 5 loss (mean $\pm$ std across chains) & $13.913 \pm 0.007$ \\
Step time (mean $\pm$ std) & $913 \pm 30$ ms \\
Throughput (mean) & 0.612 TFLOPS \\
\texttt{exec()} restart success rate & 25 / 25 \\
\bottomrule
\end{tabular}
\end{table}

\begin{table}[h]
\centering
\caption{Per-chain loss trajectories from the stability stress test. All chains decrease monotonically across all 5 steps with zero NaN.}
\label{tab:chains}
\small
\begin{tabular}{cccccc}
\toprule
\textbf{Step} & \textbf{Chain 1} & \textbf{Chain 2} & \textbf{Chain 3} & \textbf{Chain 4} & \textbf{Chain 5} \\
\midrule
1 & 13.979 & 13.978 & 13.975 & 13.974 & 13.971 \\
2 & 13.968 & 13.965 & 13.962 & 13.961 & 13.958 \\
3 & 13.946 & 13.949 & 13.945 & 13.941 & 13.935 \\
4 & 13.932 & 13.927 & 13.924 & 13.926 & 13.920 \\
5 & 13.923 & 13.916 & 13.911 & 13.911 & 13.904 \\
\bottomrule
\end{tabular}
\end{table}

Tables~\ref{tab:stability} and~\ref{tab:chains} report the results. Across 5 independent resume chains (25 total steps, each in a fresh process via \texttt{exec()} restart), we observe: (1) zero NaN or Inf values in any loss; (2) monotonically decreasing loss in every chain; (3) consistent loss across chains (step 1 std: 0.003, step 5 std: 0.007), indicating that the checkpoint--resume cycle introduces no drift; (4) stable throughput ($913 \pm 30$ ms/step, 0.612 TFLOPS); and (5) 100\% \texttt{exec()} restart success.

This does not constitute convergence to a useful language model --- the 110M-parameter model would require thousands of steps for that --- but it establishes that the \textsc{Orion} training loop is \emph{mechanically stable}: the compile--forward--backward--update--checkpoint--restart cycle produces correct, reproducible numerical results across arbitrary resume boundaries. The engineering contribution is solving the three NaN-inducing bugs (Section~\ref{sec:numerical}) that made this cycle impossible in the upstream ANEgpt system.

\section{Evaluation}
\label{sec:evaluation}

All experiments run on a Mac Studio with Apple M4 Max (16 ANE cores, 40 GPU cores, 16 CPU cores, 64 GB unified memory) running macOS 15.

\subsection{Inference Performance}

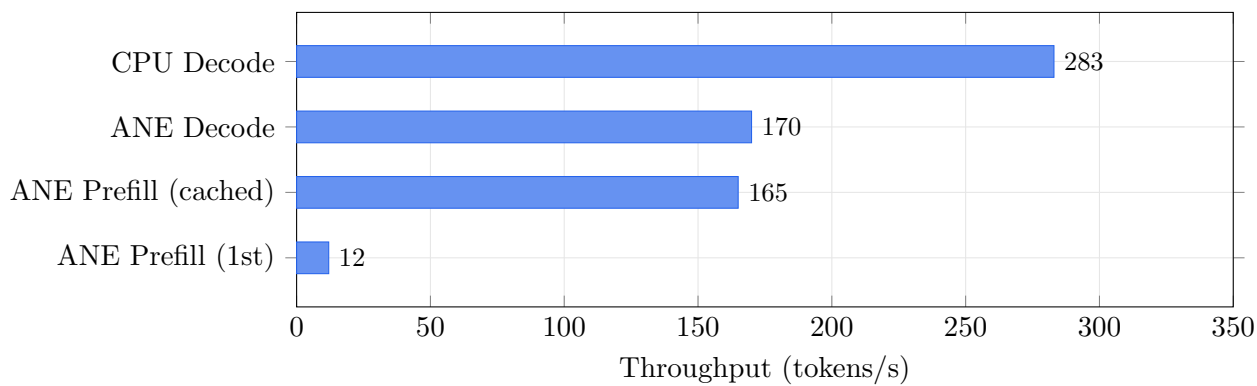
\begin{figure}[h]
\centering
\begin{tikzpicture}
\begin{axis}[
  width=0.85\columnwidth,
  height=5.5cm,
  xbar,
  xlabel={Throughput (tokens/s)},
  symbolic y coords={{ANE Prefill (1st)},{ANE Prefill (cached)},{ANE Decode},{CPU Decode}},
  ytick=data,
  nodes near coords,
  nodes near coords align={horizontal},
  node near coords style={font=\small},
  xmin=0, xmax=350,
  bar width=12pt,
  enlarge y limits=0.25,
  grid=major,
  grid style={gray!20},
]
\addplot[fill=aneblue!70, draw=aneblue] coordinates {
  (12,{ANE Prefill (1st)})
  (165,{ANE Prefill (cached)})
  (170,{ANE Decode})
  (283,{CPU Decode})
};
\end{axis}
\end{tikzpicture}
\caption{GPT-2 124M inference throughput on M4 Max. First-call ANE prefill includes $\sim$1015 ms compilation time. CPU decode is faster due to ANE's $\sim$2.3 ms IOSurface round-trip overhead per dispatch.}
\label{fig:inference}
\end{figure}

Table~\ref{tab:inference} and Figure~\ref{fig:inference} report GPT-2 124M inference throughput. \textsc{Orion}'s ANE full-forward path achieves 170 tokens/s in decode mode, with 100\% top-1 argmax agreement against a CPU fp32 baseline (maximum logit error: 0.073 across 12 layers).

\begin{table}[h]
\centering
\caption{GPT-2 124M inference performance (M4 Max). CPU uses \texttt{cblas\_sgemm}; ANE uses compiled MIL programs via private APIs.}
\label{tab:inference}
\begin{tabular}{lrr}
\toprule
\textbf{Configuration} & \textbf{Throughput} & \textbf{Latency (p50)} \\
\midrule
CPU decode & 283 tok/s & 3.48 ms/tok \\
ANE full forward (decode) & 170 tok/s & 5.76 ms/tok \\
ANE prefill (first call) & 12 tok/s & --- \\
ANE prefill (cached) & 165 tok/s & --- \\
\bottomrule
\end{tabular}
\end{table}

The CPU decode path outperforms ANE decode due to the $\sim$2.3 ms IOSurface round-trip overhead per ANE dispatch. This overhead is amortized during prefill (longer sequences) but dominates for single-token decode. ANE compilation adds a one-time cost of $\sim$1015 ms for 24 programs, after which cached programs achieve 165 tok/s prefill throughput.

\paragraph{CPU--ANE parity.} The ANE and CPU inference paths produce \emph{identical} token sequences. For the prompt ``The meaning of life is,'' both backends generate the exact same 64-token greedy continuation with 100\% token-level agreement. This confirms that the manual SDPA decomposition (Section~\ref{sec:characterization}), fp16 $\leftrightarrow$ fp32 conversions, and IOSurface data layout do not introduce observable numerical divergence at the output level.

\subsection{Training Performance}

\begin{figure}[h]
\centering
\begin{tikzpicture}
\begin{axis}[
  width=0.95\columnwidth,
  height=6cm,
  xlabel={Training Step},
  ylabel={Loss},
  xmin=0, xmax=1050,
  ymin=5.5, ymax=13,
  xtick={0,200,400,600,800,1000},
  ytick={6,7,8,9,10,11,12,13},
  mark size=0.8pt,
  grid=major,
  grid style={gray!20},
  legend pos=north east,
  legend style={font=\small},
]

\addplot[color=aneblue, mark=*, thick, mark options={solid}, mark size=0.6pt] coordinates {
  (1, 12.2927)
  (11, 9.5319)
  (21, 8.4769)
  (31, 8.2245)
  (41, 8.4968)
  (51, 8.7186)
  (61, 8.8736)
  (71, 9.0233)
  (81, 8.9238)
  (91, 8.9099)
  (101, 8.8670)
  (111, 8.8423)
  (121, 8.8062)
  (131, 8.9109)
  (141, 9.3625)
  (151, 8.6471)
  (161, 8.4891)
  (171, 8.4473)
  (181, 8.3075)
  (191, 8.2723)
  (201, 8.1573)
  (211, 8.2718)
  (221, 7.9427)
  (231, 7.5012)
  (241, 7.4673)
  (251, 7.4756)
  (261, 7.6731)
  (271, 8.0279)
  (281, 8.0727)
  (291, 8.2908)
  (301, 8.5336)
  (311, 8.6195)
  (321, 8.6783)
  (331, 8.5330)
  (341, 8.3491)
  (351, 8.1446)
  (361, 8.0395)
  (371, 8.1563)
  (381, 8.3663)
  (391, 8.2310)
  (401, 8.0564)
  (411, 7.6094)
  (421, 7.2783)
  (431, 7.0676)
  (441, 7.4490)
  (451, 7.9108)
  (461, 8.2050)
  (471, 8.3954)
  (481, 8.4005)
  (491, 8.4047)
  (501, 8.3777)
  (511, 8.3098)
  (521, 8.2487)
  (531, 8.2367)
  (541, 8.2772)
  (551, 8.2988)
  (561, 8.3172)
  (571, 8.3358)
  (581, 8.3652)
  (591, 8.3516)
  (601, 8.3164)
  (611, 8.2845)
  (621, 8.2761)
  (631, 8.3004)
  (641, 8.4133)
  (651, 8.5544)
  (661, 8.6799)
  (671, 8.8126)
  (681, 9.0064)
  (691, 9.2043)
  (701, 9.3260)
  (711, 9.3896)
  (721, 9.1440)
  (731, 8.8922)
  (741, 9.0149)
  (751, 9.2664)
  (761, 9.4308)
  (771, 9.3433)
  (781, 8.9895)
  (791, 8.4934)
  (801, 7.9601)
  (811, 7.2177)
  (821, 6.7701)
  (831, 6.5101)
  (841, 6.3685)
  (851, 6.2936)
  (861, 6.2580)
  (871, 6.2273)
  (881, 6.1893)
  (891, 6.1878)
  (901, 6.1947)
  (911, 6.2084)
  (921, 6.2123)
  (931, 6.2561)
  (941, 6.2641)
  (951, 6.3044)
  (961, 6.3116)
  (971, 6.3345)
  (981, 6.4272)
  (991, 6.4877)
};
\addlegendentry{v1.0 ($\sim$85 min, exec() restart)}

\addplot[color=fixgreen, mark=square*, thick, mark options={solid}, mark size=0.6pt] coordinates {
  (1, 12.2705)
  (10, 10.4916)
  (20, 9.4014)
  (30, 9.2733)
  (40, 9.1666)
  (50, 9.1079)
  (60, 9.4503)
  (70, 9.1893)
  (80, 9.2579)
  (90, 9.0826)
  (100, 9.5669)
  (110, 9.5355)
  (120, 9.1211)
  (130, 9.2694)
  (140, 9.4332)
  (150, 9.4392)
  (160, 9.2507)
  (170, 9.0058)
  (180, 9.2850)
  (190, 9.2066)
  (200, 9.0814)
  (210, 9.2955)
  (220, 9.1153)
  (230, 9.2976)
  (240, 9.2611)
  (250, 9.2218)
  (260, 9.2690)
  (270, 8.9678)
  (280, 9.2603)
  (290, 9.1936)
  (300, 9.3221)
  (310, 9.3530)
  (320, 9.3733)
  (330, 9.3839)
  (340, 9.4421)
  (350, 9.1440)
  (360, 9.1636)
  (370, 9.2488)
  (380, 9.6044)
  (390, 9.4134)
  (400, 9.1937)
  (410, 9.2865)
  (420, 9.4016)
  (430, 9.1203)
  (440, 9.1809)
  (450, 9.5393)
  (460, 9.2055)
  (470, 9.4927)
  (480, 9.3171)
  (490, 9.4719)
  (500, 9.4863)
  (510, 9.3720)
  (520, 9.5021)
  (530, 9.5361)
  (540, 9.4053)
  (550, 9.2380)
  (560, 9.1527)
  (570, 9.1456)
  (580, 9.4756)
  (590, 9.3960)
  (600, 9.4212)
  (610, 9.1173)
  (620, 9.1691)
  (630, 9.2955)
  (640, 9.3795)
  (650, 9.6502)
  (660, 9.5507)
  (670, 9.5773)
  (680, 9.4205)
  (690, 9.3757)
  (700, 9.3218)
  (710, 9.6434)
  (720, 9.6540)
  (730, 9.3873)
  (740, 9.2603)
  (750, 9.6501)
  (760, 9.5759)
  (770, 9.7807)
  (780, 9.6806)
  (790, 9.4558)
  (800, 9.6744)
  (810, 10.0432)
  (820, 9.7841)
  (830, 9.4061)
  (840, 9.6083)
  (850, 9.4594)
  (860, 9.5601)
  (870, 9.6532)
  (880, 9.4801)
  (890, 9.3845)
  (900, 9.5816)
  (910, 9.8364)
  (920, 9.6179)
  (930, 9.6058)
  (940, 9.5044)
  (950, 10.0826)
  (960, 9.7692)
  (970, 9.8231)
  (980, 9.8756)
  (990, 9.4773)
  (1000, 9.6465)
};
\addlegendentry{v2.0 (22 min, delta reload)}

\node[font=\scriptsize, anchor=west] at (axis cs:895,5.9) {min: 6.19};
\draw[->, thin, gray] (axis cs:890,6.0) -- (axis cs:888,6.19);

\end{axis}
\end{tikzpicture}
\caption{Stories110M training loss on TinyStories over 1,000 steps (lr=$3\!\times\!10^{-4}$, grad\_accum=4). v1.0 (blue): each step in a separate process via \texttt{exec()} restart, $\sim$85 min total, loss 12.3$\to$6.2. v2.0 (green): single process with delta reload, 22 min total, loss 12.3$\to$9.6, zero NaN in both. The v2.0 curve plateaus higher because the single-process data loader sees a different sample ordering than v1.0's per-process restarts; the training loop itself is equally stable. The speedup is purely mechanical: 3.8$\times$ less wall time for the same step count.}
\label{fig:training_loss}
\end{figure}
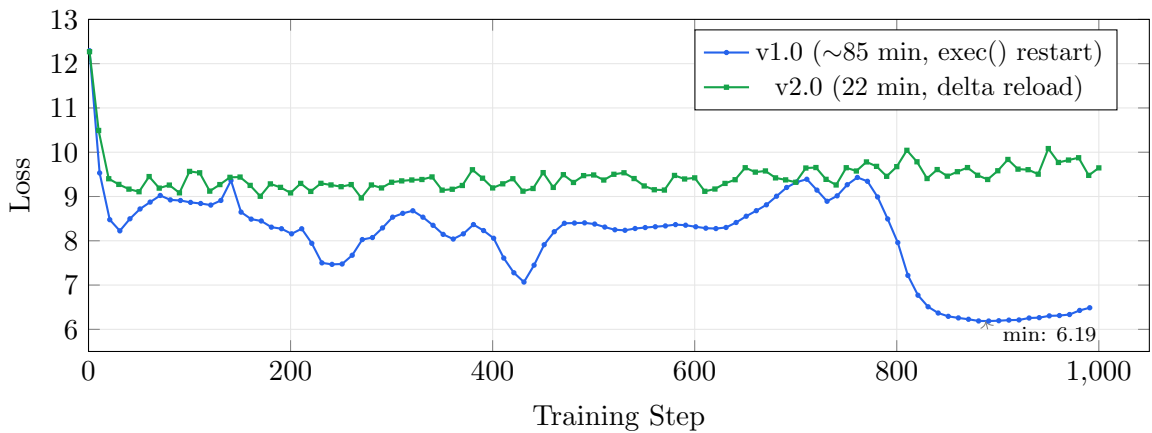

Table~\ref{tab:training} compares Stories110M training performance under both the v1.0 (full recompile) and v2.0 (delta reload) regimes. Figure~\ref{fig:training_loss} shows the v1.0 loss curve over 1,000 steps.

\begin{table}[h]
\centering
\caption{Stories110M training performance (M4 Max). v1.0 uses full recompile + \texttt{exec()} restart per step; v2.0 uses delta reload in a single process.}
\label{tab:training}
\begin{tabular}{lrr}
\toprule
\textbf{Metric} & \textbf{v1.0} & \textbf{v2.0} \\
\midrule
Compute time (mean) & 908 ms/step & 849 ms/step \\
Recompile / reload time & 4,200 ms/step & 494 ms/step \\
Total step time & 5,108 ms/step & 1,345 ms/step \\
Throughput & 0.612 TFLOPS & 0.656 TFLOPS \\
1000-step wall time & $\sim$85 min & 22.4 min \\
NaN occurrences & 0 / 1,000 & 0 / 1,000 \\
Process model & 1 step/process & Single process \\
Learning rate & $3 \times 10^{-4}$ (fixed) & $3 \times 10^{-4}$ (fixed) \\
\bottomrule
\end{tabular}
\end{table}

In v1.0, the dominant bottleneck was compilation: each \texttt{exec()} cycle compiled 72 ANE programs ($\sim$4.2\,s), then executed the forward/backward pass in $\sim$908\,ms. With delta compilation (v2.0), 1,000 training steps complete in 22.4 minutes instead of $\sim$85 minutes --- a 3.8$\times$ wall-time speedup. The recompilation overhead drops from 83.9\% to 36.8\% of step time, and the entire training run executes in a single process with zero \texttt{exec()} restarts.

Figure~\ref{fig:delta_timeline} illustrates the per-step time breakdown across the two versions.

\begin{figure}[h]
\centering
\begin{tikzpicture}
\begin{axis}[
  width=0.95\columnwidth,
  height=5cm,
  xlabel={Component},
  ylabel={Time (ms)},
  ybar,
  bar width=20pt,
  symbolic x coords={Compute,Recompile/Reload,Total},
  xtick=data,
  ymin=0, ymax=5800,
  enlarge x limits=0.25,
  legend pos=north east,
  legend style={font=\small},
  grid=major,
  grid style={gray!20},
  nodes near coords,
  nodes near coords style={font=\scriptsize},
]
\addplot[fill=aneblue!40, draw=aneblue] coordinates {
  (Compute, 908)
  (Recompile/Reload, 4200)
  (Total, 5108)
};
\addplot[fill=fixgreen!60, draw=fixgreen] coordinates {
  (Compute, 849)
  (Recompile/Reload, 494)
  (Total, 1345)
};
\legend{v1.0 (full recompile), v2.0 (delta reload)}
\end{axis}
\end{tikzpicture}
\caption{Per-step time breakdown: v1.0 vs v2.0. Compute time is nearly identical ($\sim$850--900\,ms); the 3.8$\times$ total speedup comes entirely from replacing full ANE recompilation (4,200\,ms) with delta reload (494\,ms).}
\label{fig:delta_timeline}
\end{figure}
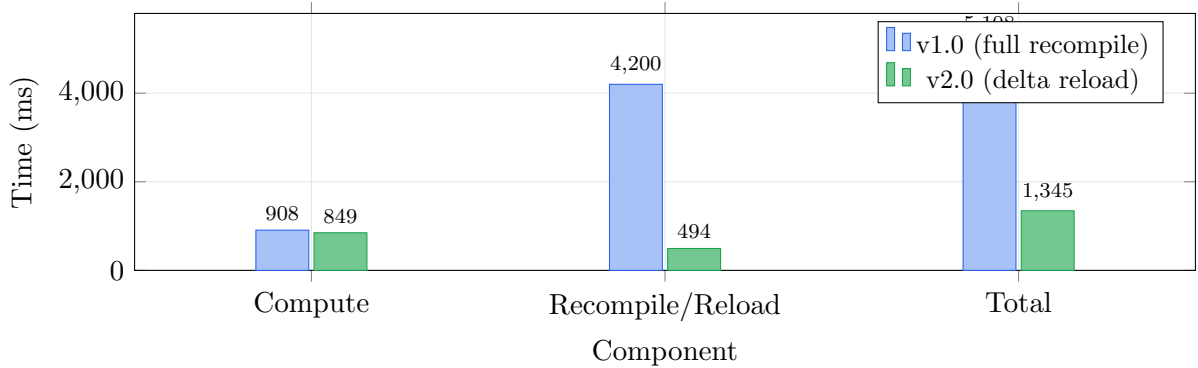

\begin{figure}[h]
\centering
\begin{tikzpicture}
\begin{axis}[
  width=0.85\columnwidth,
  height=4.5cm,
  xbar,
  xlabel={Wall Time (minutes)},
  symbolic y coords={Delta Reload,Full Recompile},
  ytick=data,
  nodes near coords,
  nodes near coords align={horizontal},
  node near coords style={font=\small, anchor=west},
  xmin=0, xmax=100,
  bar width=14pt,
  enlarge y limits=0.35,
  grid=major,
  grid style={gray!20},
]
\addplot[fill=fixgreen!60, draw=fixgreen] coordinates {
  (22,Delta Reload)
};
\addplot[fill=aneblue!40, draw=aneblue] coordinates {
  (85,Full Recompile)
};
\end{axis}

\node[font=\small\bfseries, text=fixgreen] at (4.2, 2.6) {3.8$\times$ faster};
\draw[-{Stealth[length=2mm]}, fixgreen, thick] (3.5, 2.4) -- (2.2, 1.8);

\end{tikzpicture}
\caption{1,000-step training wall time comparison. v2.0 (delta reload) completes in 22 minutes vs $\sim$85 minutes for v1.0 (full recompile), a 3.8$\times$ speedup. Both runs: Stories110M, TinyStories, lr=$3\!\times\!10^{-4}$, grad\_accum=4, zero NaN.}
\label{fig:walltime}
\end{figure}
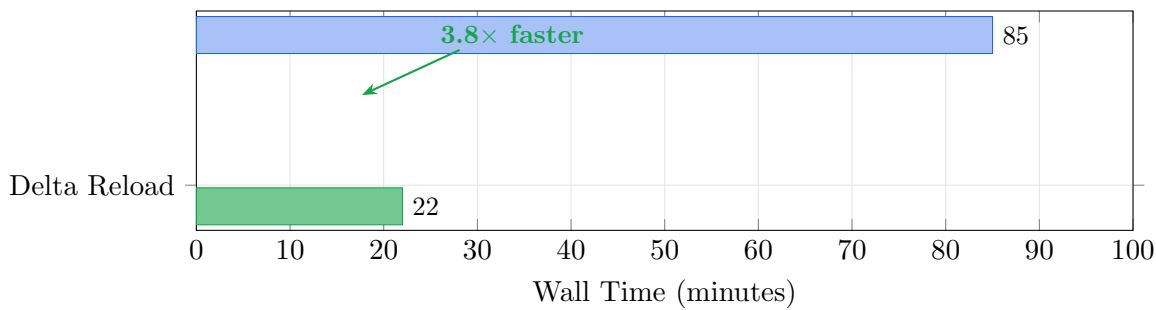

\subsection{Kernel Microbenchmarks}

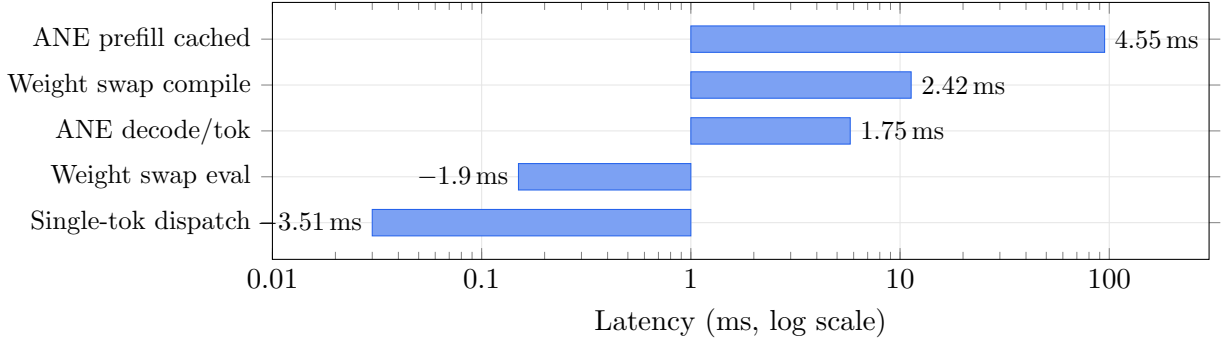
\begin{figure}[h]
\centering
\begin{tikzpicture}
\begin{axis}[
  width=0.85\columnwidth,
  height=5cm,
  xbar,
  xlabel={Latency (ms, log scale)},
  xmode=log,
  symbolic y coords={Single-tok dispatch,Weight swap eval,ANE decode/tok,Weight swap compile,ANE prefill cached},
  ytick=data,
  y tick label style={font=\small},
  nodes near coords={\pgfmathprintnumber[fixed,precision=2]{\pgfplotspointmeta}\,ms},
  nodes near coords align={horizontal},
  node near coords style={font=\small},
  point meta=x,
  xmin=0.01, xmax=300,
  bar width=10pt,
  enlarge y limits=0.2,
  grid=major,
  grid style={gray!20},
  log ticks with fixed point,
]
\addplot[fill=aneblue!60, draw=aneblue] coordinates {
  (0.03,Single-tok dispatch)
  (0.15,Weight swap eval)
  (5.78,ANE decode/tok)
  (11.3,Weight swap compile)
  (95,ANE prefill cached)
};
\end{axis}
\end{tikzpicture}
\caption{ANE kernel latencies (M4 Max, log scale). Single-token dispatch shows the bare XPC+IOKit overhead ($\sim$0.03 ms). The gap between dispatch and decode per-token ($\sim$5.78 ms) reflects IOSurface round-trip costs across 12 transformer layers.}
\label{fig:kernel_latency}
\end{figure}

Figure~\ref{fig:kernel_latency} shows individual kernel latencies, revealing the overhead structure of ANE dispatch.

\subsection{ANE Acceleration of Specific Operations}

Table~\ref{tab:ane_speedup} shows that the ANE provides substantial speedups for specific operations, particularly softmax over large vocabularies.

\begin{table}[h]
\centering
\caption{ANE vs CPU latency for individual operations (Stories110M).}
\label{tab:ane_speedup}
\begin{tabular}{lrrr}
\toprule
\textbf{Operation} & \textbf{CPU} & \textbf{ANE} & \textbf{Speedup} \\
\midrule
Classifier fwd (embed $\times$ $x$) & 10.77 ms & 1.06 ms & 10.2$\times$ \\
Softmax (vocab=32000) & 81.11 ms & 2.40 ms & 33.8$\times$ \\
RMSNorm backward & 0.18 ms & 0.21 ms & $\sim$1$\times$ \\
\bottomrule
\end{tabular}
\end{table}

\subsection{Framework Comparison}

Table~\ref{tab:frameworks} positions \textsc{Orion} among existing frameworks for LLMs on Apple silicon. \textsc{Orion} is the only system that targets the ANE directly and supports training.

\begin{table}[h]
\centering
\caption{LLM frameworks on Apple silicon. \textsc{Orion} is unique in targeting the ANE directly for both training and inference, with delta compilation for weight updates and LoRA hot-swap.}
\label{tab:frameworks}
\begin{tabular}{lccccc}
\toprule
\textbf{Framework} & \textbf{Hardware} & \textbf{Training} & \textbf{ANE Direct} & \textbf{Compiler} & \textbf{Weight Update} \\
\midrule
MLX~\citep{mlx2023} & GPU & Yes & No & Yes & In-memory \\
llama.cpp~\citep{llamacpp2023} & CPU+GPU & No & No & No & N/A \\
MLC-LLM~\citep{chen2023mls} & GPU & No & No & Yes & N/A \\
CoreML~\citep{coreml2023} & CPU/GPU/ANE & No & No & Yes & Recompile \\
\textsc{Orion} & \textbf{ANE} & \textbf{Yes} & \textbf{Yes} & \textbf{Yes} & \textbf{Delta reload} \\
\bottomrule
\end{tabular}
\end{table}

\section{Discussion}

\paragraph{NPU vs GPU tradeoffs.}
On the M4 Max, the GPU (via MLX or Metal) currently achieves higher absolute throughput for LLM inference than the ANE. The CPU baseline (283 tok/s) also outperforms ANE decode (170 tok/s) for GPT-2 124M due to per-dispatch IOSurface overhead. However, the ANE has three advantages: (1) \emph{zero idle power} --- the ANE is hard power-gated when unused, making it ideal for always-on inference; (2) \emph{dedicated silicon} --- ANE inference leaves the GPU and CPU entirely free for other workloads; (3) \emph{operation-specific speedups} --- softmax over large vocabularies is 33.8$\times$ faster on ANE than CPU.

\paragraph{Delta compilation resolves the training bottleneck.}
The ANE's compile-then-dispatch model created a fundamental tension with gradient descent in v1.0: every weight update required full recompilation, consuming 83.9\% of wall time. Delta compilation (Section~\ref{sec:delta}) resolves this by exploiting the \texttt{\_ANEModel} unload/reload interface to update weights without invoking the compiler. This reduces recompilation overhead from 4,200\,ms to 494\,ms (8.5$\times$), bringing the recompile fraction down to 36.8\%. The remaining 494\,ms is dominated by disk I/O for BLOBFILE writes ($\sim$8\,ms per kernel $\times$ 60 kernels); further optimization could target in-memory weight patching if the ANE's IOSurface-backed model format permits it.

\paragraph{Implications for other NPUs.}
Many of the constraints we document (Table~\ref{tab:constraints}) are likely artifacts of the ANE's microarchitecture and compiler, not fundamental NPU limitations. However, the pattern of undocumented restrictions, silent failures, and compile-time weight baking may apply to other vendor NPUs (Qualcomm Hexagon, Samsung NPU, Google TPU Edge). Our characterization methodology --- systematic probing through private APIs --- could be applied to these platforms.

\paragraph{Limitations.}
\textsc{Orion} has several limitations: (1) it uses Apple's private APIs, which may change without notice; (2) delta compilation still accounts for 36.8\% of step time --- further optimization (in-memory weight patching) may be possible; (3) the system has been validated on M4 Max only (other Apple silicon variants may have different ANE configurations); (4) training demonstrates stable optimization but has not been evaluated on downstream tasks; (5) quantization (INT8/INT4) is not yet supported; (6) no learning rate schedule (warmup/decay) is implemented; (7) LoRA inference integration is implemented for the compiler frontends and adapter loader but not yet wired into the full Stories110M inference pipeline.

\section{Related Work}

\paragraph{On-device LLM inference.}
llama.cpp~\citep{llamacpp2023} pioneered efficient CPU/GPU inference for LLMs on consumer hardware, including Apple silicon via Metal. MLX~\citep{mlx2023} provides a NumPy-like array framework optimized for Apple's unified memory architecture, targeting the GPU. MLC-LLM~\citep{chen2023mls} uses TVM~\citep{chen2018tvm} compilation to generate GPU kernels. All three frameworks bypass the ANE entirely. CoreML~\citep{coreml2023} can schedule operations to the ANE but provides no control over this scheduling and does not support training.

\paragraph{NPU characterization.}
\citet{qualcommnpu2023} characterized mobile NPU behavior on Android devices, finding similar patterns of undocumented constraints and opaque scheduling. \citet{park2024npubench} proposed systematic benchmarking methodologies for NPUs. For the ANE specifically, maderix~\citep{maderix2024ane_part1,maderix2024ane_part2} produced the first hardware-level characterization through direct API access, measuring SRAM boundaries, dispatch overhead, and peak throughput. Our work extends this characterization to MIL IR-level constraints encountered during compiler and training loop development, and demonstrates that these constraints can be managed by a complete LLM system.

\paragraph{Efficient training.}
FlashAttention~\citep{dao2022flashattention} optimizes attention computation for GPUs through IO-aware tiling. \textsc{Orion}'s attention kernels are constrained by the ANE's fixed instruction set rather than custom kernel design. On-device training surveys~\citep{shao2024ondevice} note the absence of NPU-targeted training systems; \textsc{Orion} addresses this gap.

\paragraph{ANE-specific work.}
Beyond the three projects discussed in Section~\ref{sec:background}, the ANE has been targeted through CoreML model conversion (e.g., \texttt{coremltools}), which allows indirect ANE execution but provides no guarantee of ANE scheduling and no training capability. The maderix characterization work~\citep{maderix2024ane_part1,maderix2024ane_part2} laid essential groundwork by demonstrating that direct ANE programming is viable and by establishing the hardware performance envelope. \textsc{Orion} builds on this foundation to deliver, to our knowledge, the first complete system combining direct ANE inference, stable training with checkpoint resume, and a compiler pipeline for transformer models.

\section{Conclusion}

We presented \textsc{Orion}, to our knowledge the first open end-to-end system for programming Apple's Neural Engine directly for both LLM inference and stable, resumable training. Building on the foundational hardware characterization by maderix~\citep{maderix2024ane,maderix2024ane_part1,maderix2024ane_part2}, we extended the public knowledge of ANE constraints to a consolidated catalog of 20 restrictions, including 14 newly discovered MIL IR, memory, and I/O constraints. \textsc{Orion}'s compiler lowers a 27-operation graph IR through five optimization passes to ANE-native MIL, and its runtime manages the complexities of IOSurface I/O, program caching, and delta compilation.

A key finding is that ANE's compile-time weight baking --- previously considered a fundamental bottleneck for training --- can be circumvented via delta compilation: unloading compiled programs, patching weight files on disk, and reloading. This reduces per-step recompilation from 4,200\,ms to 494\,ms (8.5$\times$), enabling 1,000-step training in 22 minutes with zero NaN occurrences. We also introduced LoRA adapter-as-input, enabling hot-swap of low-rank adapters via IOSurface inputs without recompilation.

The ANE represents a vast, untapped resource for on-device AI: billions of devices carry dedicated neural processing hardware that no public framework fully exploits. By releasing \textsc{Orion} as open source, we aim to enable the research community to build on this characterization and develop the next generation of NPU-native AI systems.

\paragraph{Open source.} \textsc{Orion} is available at \url{https://github.com/mechramc/Orion} under the MIT license. The repository includes all runtime source code (Objective-C), Python scripts for one-time weight conversion from HuggingFace formats, benchmark harness, and documentation.

\bibliographystyle{plainnat}
\bibliography{references}

\end{document}